\documentclass[10pt,conference]{IEEEtran}
\usepackage{graphicx} 
\usepackage{acronym}
\usepackage{cite}
\usepackage{amsmath}
\usepackage{amssymb}
\usepackage{hyperref}
\usepackage{csquotes}
\usepackage{listings}
\usepackage{multirow}
\usepackage{booktabs}
\usepackage{xcolor}
\usepackage{microtype}

\title{Structural Generalization in\\ Autonomous Cyber Incident Response with Message-Passing Neural Networks and\\ Reinforcement Learning}

\author{\IEEEauthorblockN{1\textsuperscript{st} Jakob Nyberg}
\IEEEauthorblockA{\textit{Division of Network and Systems Engineering} \\
\textit{KTH Royal Institute of Technology}\\
Stockholm, Sweden \\
jaknyb (at) kth.se}
\and
\IEEEauthorblockN{2\textsuperscript{nd} Pontus Johnson}
\IEEEauthorblockA{\textit{Division of Network and Systems Engineering} \\
\textit{KTH Royal Institute of Technology}\\
Stockholm, Sweden \\
pontusj (at) kth.se}
}

\acrodef{gnn}[GNN]{graph neural network}
\acrodef{mpnn}[MPNN]{message-passing neural network}
\acrodef{cnn}[CNN]{convolutional neural network}
\acrodef{mlp}[MLP]{multi-layer perceptron}
\acrodef{cage}[CAGE]{Cyber Autonomy Gym for Experimentation}
\acrodef{cyborg}[CybORG]{Cyber Operations Research Gym}
\acrodef{ids}[IDS]{intrusion detection system}
\acrodef{nids}[NIDS]{network intrusion detection system}
\acrodef{ppo}[PPO]{proximal policy optimization}
\acrodef{gae}[GAE]{generalized advantage estimation}
\acrodef{rl}[RL]{reinforcement learning}
\acrodef{ttcp}[TTCP]{Technical Cooperation Program}
\acrodef{wl}[WL]{Weisfehler-Lehman}
\acrodef{mdp}[MDP]{Markov decision process}
\newacroindefinite{mdp}{an}{a}
\newacroindefinite{mpnn}{an}{a}
\newacroindefinite{mlp}{an}{a}
\newcommand{\vertex}{\mathcal{V}}

\begin{document}

\maketitle





\begin{abstract}
    We believe that agents for automated incident response 
    based on machine learning need to handle changes in network structure.
    Computer networks are dynamic, and can naturally change in structure over time.
    Retraining agents for small network changes costs time and energy. 
    We attempt to address this issue with an existing method of relational agent learning,
    where the relations between objects are assumed to remain consistent across problem instances.
    The state of the computer network is represented as a relational graph and encoded through a message passing neural network. 
    The message passing neural network and an agent policy using the encoding are optimized end-to-end using reinforcement learning.
    We evaluate the approach on the second instance of the Cyber Autonomy Gym for Experimentation (CAGE~2), 
    a cyber incident simulator that simulates attacks on an enterprise network.
    We create variants of the original network with different numbers of hosts 
    and agents are tested without additional training on them. 
    Our results show that agents using relational information are able to find solutions despite 
    changes to the network, and can perform optimally in some instances.
    Agents using the default vector state representation perform better, 
    but need to be specially trained on each network variant, demonstrating a trade-off between specialization and generalization.
\end{abstract}



\section{Introduction}

We focus on the automation of cyber security incident response, 
where we imagine an automated \emph{agent} that takes actions and modifies the computer network to address ongoing incidents. 
An example scenario is ransomware spreading in a corporate network, where action is required to prevent further damage.
Following previous work in this domain~\cite{applebaumtabular, 5428673, collyer2022acd, cyborgrewardshaping, hammar2020finding, wolk2022cage}, we frame incident response as a stochastic decision problem, 
or a game between agents defending the network and agents that attempt to compromise it. 
A common method used to find a solution to this class of problem is \ac{rl}~\cite{sutton2018reinforcement}, 
which has been applied in the context of incident response by several authors~\cite{9596578, ADAWADKAR2022105116}.
\Ac{rl} assumes the dynamics of the problem is not known, which is common for real-world problems, 
but that the agent can receive observations of the problem state and affect the state through a set of actions.

Computer networks are often dynamic in structure, with hosts connecting and disconnecting 
over time. We consider it an important feature that a network management agent can handle changes to the network with no additional training, so called \emph{zero-shot generalization}.
There are several proposed methods for facilitating zero-shot generalization in reinforcement learning~\cite{DBLP:journals/jair/KirkZGR23}.
One approach is \emph{relational} reinforcement learning, where knowledge about relations between objects in a domain 
is assumed to be transferable across problem instances~\cite{DBLP:journals/ml/DzeroskiRD01}. 

We use an approach for relational reinforcement learning proposed by \cite{janisch2023symbolic} and adapt it to a network security problem. The state of the computer network is represented as a relational graph, which is encoded with a \acf{mpnn}. An agent then uses this encoding to take actions on objects represented in the graph. We also evaluate a modified version of the approach where the agent only incorporates information from the local neighborhood of the graph.
We evaluate the \ac{mpnn} agent on the second iteration of the \ac{cage}, a
two-player cyber security incident simulator. Previous work has noted that
agents trained in this environment have issues with structural generalization~\cite{wolk2022cage}. We create a number of variants of the original
network topology used in \ac{cage} 2 and use these to test the generalization capabilities of agents. 
Our results demonstrate that an agent which encodes the network state using \iac{mpnn} can generalize with no additional training on these variants, something an agent using a \ac{mlp} encoder and vector representations can not. 

\section{Background}



\begin{figure}
    \centering
    \includegraphics[width=0.45\textwidth]{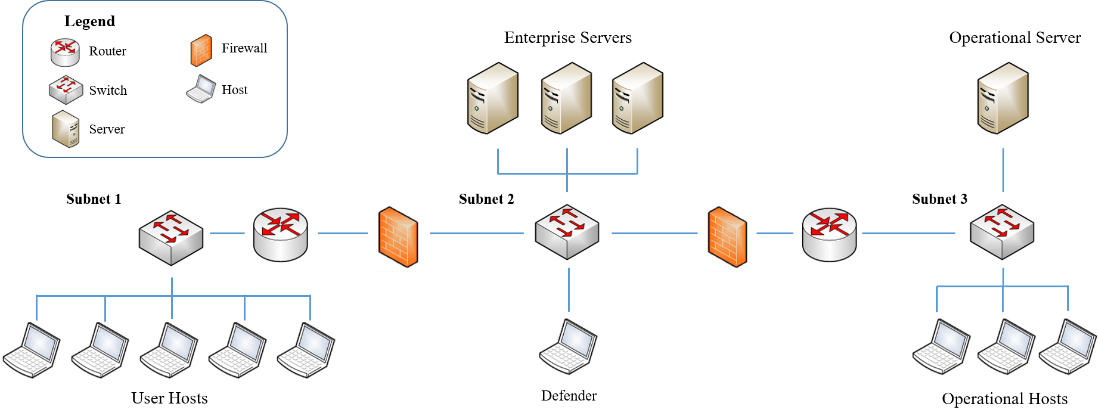}
    \caption{Image of computer network simulated in CAGE 2 presented on the developers GitHub repository. During a match, the red team starts from an user host and attempts to reach the operational server. This can be achieved by first compromising one of the enterprise servers, which can communicate with the operational machines.}\label{fig:cage2}
\end{figure}

\subsection{CAGE and CybORG}\label{sec:cyborg}

The \acf{cage} is a collection of four cyber security scenario simulators developed by the \ac{ttcp}. All were released as part of public competitions to facilitate the development of automated cyber operation agents. The \ac{cage} scenarios share a common underlying simulator infrastructure, the \ac{cyborg}~\cite{cyborg_acd_2021}. Although initially described as a simulator where scenarios could be replicated in an emulator of virtual machines, the publicly released portions of \ac{cyborg} consists of only the simulator. The developers have confirmed that development of the emulator has ceased.
The first two scenarios are set in a corporate network and use the same network topology, with the second scenario adding more actions for the red and blue agents. The third \ac{cage} scenario consists of a network of aerial drones under attack. The fourth and most recent scenario returns to the network defense domain, but in a multiagent context. We focus on the second scenario, which we will refer to it as \ac{cage} 2\footnote{\href{https://github.com/cage-challenge/cage-challenge-2}{www.github.com/cage-challenge/cage-challenge-2}}.

\begin{figure}[!htpb] \centering
\includegraphics[width=0.48\textwidth]{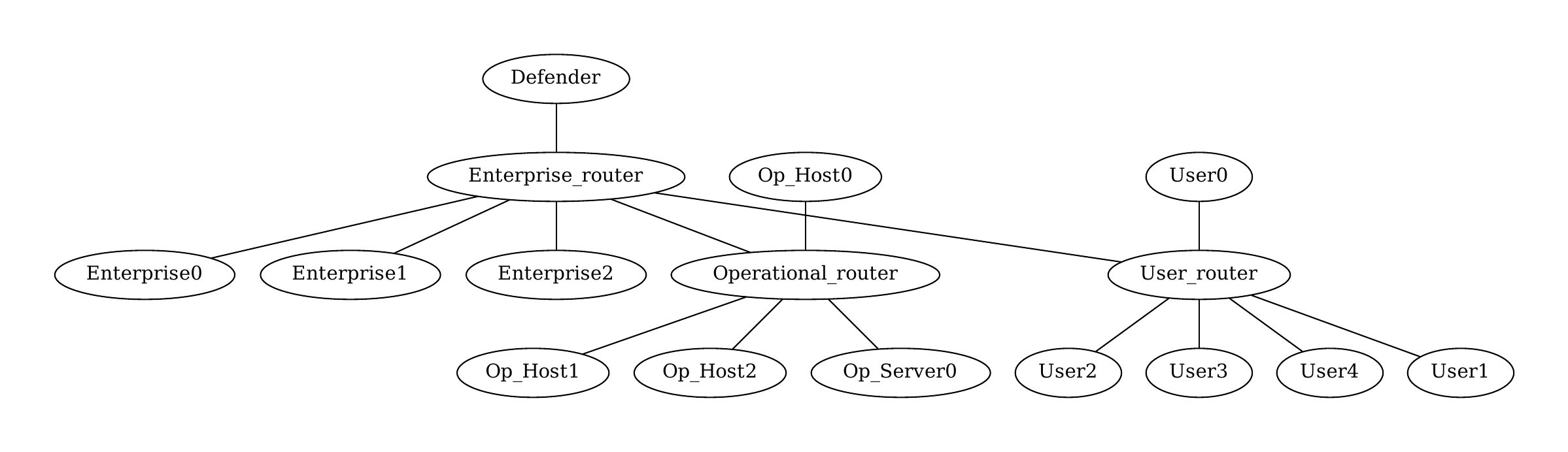}
\caption{
    Graph representation of network structure used in CAGE 2.
 }\label{fig:network}
\end{figure}

The scenario premise of CAGE 2 is that a computer network in a manufacturing plant is under attack by a hostile entity. There are two teams acting within the network:
A \enquote{red} team, consisting of an automated agent, attempts to gain control of machines in the network, with the ultimate goal of reaching the operational server; A \enquote{blue} team of network operators attempts to stop the red team from compromising hosts in the network. The network consists of 13 hosts divided into three subnets, separated by firewalls that only allow traffic between certain subnets. An image from the CAGE 2 GitHub repository showing the computer network is presented in \ref{fig:cage2}. A graph representation of the network is shown in \autoref{fig:network}, with host names as nodes and edges representing connections between them. A third, \enquote{green}, team is included in the scenario description, representing normal users causing false alerts. However, it is not used in the challenge evaluation, nor in previous work as far as we are aware. 

The blue team can act in the network using 13 commands. Ten commands take a host descriptor as a singular parameter, and two actions use no parameters. These are the commands \enquote{Sleep} and \enquote{Monitor}, which function as \emph{NoOp} actions.
The \enquote{Remove} command will remove red team access from a specified host. 
The red team can gain two levels of authority on a host. If the red team has reached the second level, \enquote{Remove} will not work
and access can only be removed by the \enquote{Restore} command. The command
\enquote{Analyze} returns a list of processes and 
files on a specified host. This can inform the blue team that the red team has gained access to the host.
The blue team can use eight \enquote{Decoy} commands\footnote{There is a discrepancy between the code and the documentation. The code contains eight decoys, but the documentation only lists seven. The \enquote{Sleep} command is not listed either.}. 
These create processes on a specified host that the red team will always fail to exploit.

The observed state provided to both teams is a mapping from host names to lists of simulated processes, network connections and files.
CAGE 2 also provides a wrapper function that encodes the mapping into a single vector. 
The wrapper function applies a set of heuristics to the
host data, and provides two categorical values per host. One indicates a
type of activity that has been observed on the host, which is based on port usage. 
The other shows what level of access privileges the red agent has obtained on the host.
This is based on the number of open connections, and the list of files on the host.
The wrapper function encodes the categorical values to binary, 
making the state vector \(4\cdot13=52\) elements long.
In addition to simplifying the observation space, the wrapper 
flattens the parametric action space. This gives the blue team \(2 + (13 \cdot 11) = 145\) actions. All related works we are aware of have used 
the wrapped observation and action spaces for blue team agents.
There is no defined terminal state for the simulation. The evaluation used for the challenge
truncates each round at one of 30, 50 or 100 steps. 
During an ongoing game, the red team can continually keep attempting to compromise hosts as they can never 
be fully removed from the network.
\ac{cage} 2 includes two heuristic agents to act as the red team, named \enquote{B-Line} and \enquote{Meander}. \enquote{Meander} operates similar to a breadth-first search, where it attempts to gain access to all machines in a subnet before moving on to the next. \enquote{B-line} selects hosts from a predefined list of hosts, which constitute the shortest path between the attacker starting point and the operational server.

Scoring for the blue team is based on both red team actions and their own actions. When the red team gains access to a system, the blue team
receives a penalty. The size of the penalty is proportional to the importance of the host to the organization, with the operational server being the most important.
The blue team also receives a penalty each time the \enquote{Restore} action is performed on a host.
As part of the challenge, \ac{cage} 2 uses a standardized evaluation scheme. Agents are evaluated against both the B-Line and Meander agents on 1000 episodes with lengths of 30, 50 and 100 steps. Average rewards are calculated for each pairing, and a total score is calculated as the sum of average rewards. The highest possible score for is 0.




\subsection{Learning from Graphs}\label{sec:gnn}

Graphs are data structures that represent relationships or interactions between
objects or entities. In a simple formulation, a graph consists of a set of
\emph{nodes} \(\vertex = \left\{ v_1, \ldots, v_n \right\}\), 
representing objects, and \emph{edges} \(\mathcal{E} \subseteq \left\{ (u, v)\ |\ (u, v) \in \vertex^2  \right\}\) representing relations between objects. Many functions and algorithms have been proposed to analyze aspects of graph-structured data~\cite{graphlearning}. We focus on \emph{message-passing} algorithms, where node representations are at every iteration updated using an aggregation of \emph{messages} from their immediate neighborhood. A general message passing update rule is
\begin{align} h_v^{l} = \text{UPDATE}({\text{AGGREGATE}(v)}, h_v^{l-1})
\label{eq:mp}
\end{align}
where \(\text{AGGREGATE}\) is a permutation invariant aggregation
function, and \(\text{UPDATE}\) is a function of the aggregated neighborhood of \(v\) and the node feature at step \(l\).
A graph representations can be obtained through an aggregation of the node
representations. As long as all aggregations are permutation invariant, 
message passing algorithms create representations of the graph that are permutation invariant and equivariant~\cite[Ch. 5]{graphlearning}. 


The update and aggregation functions of the message-passing function in \autoref{eq:mp} can be implemented using neural networks, forming as class of methods called \acf{mpnn}. \acp{mpnn} shares parameters across nodes, meaning that although the size of an input graph may change, the number of parameters in the model does not necessarily need to. There are several proposed architectures of \ac{mpnn}, which use different schemes of aggregations and updates. These have different abilities to represent and discriminate graph structures, and some will fail to discriminate certain graph structures~\cite{gin}. 

Representations produced by message passing are dependent on the number of iterations the algorithm runs, which is equivalent to the number of layers in \iac{mpnn}. This dictates the size of the \emph{receptive field} of each node or, put in another way, how far information travel between nodes in the graph.
With too few steps, node features may not be expressive enough
to be distinguished. On the other hand, as the number of message-passing steps increase, 
node representations may become homogenized across the graph, a concept referred to as \emph{over-smoothing}.
The optimal number of message passing iterations is dependent on the graph and the downstream task. 

The term \ac{mpnn} is sometimes used interchangeably with \ac{gnn}. We prefer to use the term \ac{mpnn} as not all neural networks applied on graphs use message-passing, making \ac{gnn} a broader descriptor than the methods we use require.

\subsection{Reinforcement Learning}

\newcommand{\state}{S} \newcommand{\action}{A}

Reinforcement learning is a field of machine learning for solving decision and control problems~\cite{sutton2018reinforcement}. Problems solved with reinforcement learning can usually be expressed as \iac{mdp}. \Iac{mdp} can be described as a tuple of states, actions, rewards and state transition probabilities. The reward function \(R: A \times S \rightarrow \mathbb{R}\) associates each state and action pair with a real value. For a finite \ac{mdp}, the goal of an agent is to maximize the expected total reward \(G=\sum_t^T R_t\), where \(T\) is the step the \ac{mdp} terminates at.

In each state an agent decides on an action using a \emph{policy}, 
a function from a state to an action or distribution over actions, \(\pi : S \rightarrow \Delta(A)\). 
The \emph{value} of a policy \(\pi\) is denoted as \(V^{\pi}\), 
and is the expected total reward in a state \(s\) when following the policy \(\pi\). 
An optimal policy \(\pi^*\) is associated with the optimal value function \(V^* :=\max_\pi V_\pi(s)\). Reinforcement learning methods attempt to approximate the optimal policy or value function through continuous interaction with the problem. Policy gradient reinforcement learning methods optimize the policy function towards the optimal policy through gradient descent. Actor-critic methods extend this method by also estimating the value function as part of the optimization process. 


For some problems, decisions can naturally be factorized into a series or
set of action variables. This includes the \ac{cage} 2 environment described in
Section~\ref{sec:cyborg}, where actions are separated into hosts and commands. 
Given a problem with a joint action \(\textbf{a}=(a_1\ldots,a_i)\), the probability of taking \(\textbf{a}\) in the state \(\state\) can be expressed as a product of conditional probabilities,
\begin{align} \pi(\mathbf{a} | \state) = \prod_i^{|\mathbf{a}|} \pi_i(a_i | \state,
	a_1,\ldots, a_{i-1})\end{align}
Similar forms of policy factorization has been used in previous works on reinforcement learning~\cite{vinyals2017starcraft, hammar2020finding, janisch2023symbolic}. 

\subsection{Symbolic Relational Deep Reinforcement Learning}\label{sec:srdrl}

We use an architecture for relational learning named Symbolic Relational Deep
Reinforcement Learning (SR-DRL)~\cite{janisch2023symbolic}. SR-DRL is intended
for problem domains that can be represented using objects and actions are taken
on those objects. Objects in the problem and their relations are represented as a relational
graph, 
and \iac{mpnn} is used to encode the graph as input to a policy function. Policy factorization is used to separate joint actions into conditional parameters for additional flexibility in representation.
SR-DRL generates a whole-graph representation in addition to node representations.
The graph representation is updated sequentially after each message-passing step, incorporating the previous representation in the next iteration. 
The \ac{mpnn} and policy function are trained end-to-end using reinforcement learning.
An action-critic scheme is used when training agents, where
the value function estimator is implemented as a neural network taking the graph representation as input.



\section{Implementation of MPNN Agents on CAGE 2}

To apply the SR-DRL approach on \ac{cage} 2, we reshape the vector observation space into a relational graph. We also modify the action space. An agent has to make two decisions at each turn: a command to run and a host to run it on. This is reformulated as the agent selecting a node in the graph, and then an action based on the node representation. We use \ac{ppo} to learn a policy function, but any policy gradient method, such as A2C can be used instead.
Code use for training and evaluating agents is available on GitHub\footnote{\href{https://github.com/kasanari/incident-response-rl-gnn}{www.github.com/kasanari/incident-response-rl-gnn}}



As described in Section~\ref{sec:srdrl}, SR-DRL incorporates
the full graph representation as part of the local node representations~\cite{janisch2023symbolic}.
We hypothesize that including the graph representation in the state
is unnecessary for problems where reasoning only requires information
from the local neighborhood of a node.
To test this, we evaluate a modified agent that removes the graph component of the policy
input alongside the original version used for SR-DRL.
The resulting agent only incorporates the graph representation to estimate the
value function during training.
During inference, the agent only uses information from a \(k\)-hop neighborhood to act, where \(k\) is the number of message-passing steps. Another change that comes from using only local information is the policy factorization. The implementation of SR-DRL factorizes the policy such that the agent first selects an action to take, based on the global graph representation, and then what object to act on.
In order to only use the local context, we flip this ordering.
The agent thus first selects the object to act on, and then what command to execute on it.

\subsection{Local Message-Passing Scheme}\label{sec:message-passing}

\newcommand{\command}{\mathcal{C}}
\newcommand{\nodepolicy}{\pi_1(\vertex{} = v_i|\nodeembed{})}
\newcommand{\condpi}{\pi_2(\command{} = c_j|h_i^l, \vertex=v_i)}
\newcommand{\nodeembed}{\mathbf{H}^l}
\newcommand{\graphembed}{g^l} \newcommand{\param}{\mathbf{W}}
\newcommand{\basis}{e_{i}} \newcommand{\nn}{\varphi}
\newcommand{\nnodes}{|\vertex|}
The local message passing scheme is largely unchanged from the original implementation~\cite{janisch2023symbolic}, albeit with the graph representation removed from the node update.
The update to the representation \(h\) for a node \(v\) at layer \(l\) is thus defined as
\begin{align} h_v^{l} = h_v^{l-1} + \phi_{agg}((v, v_{msg})),
\end{align}
where
\begin{align} v_{msg} = \max_{u \in \mathcal{N}(n)}
  \phi_{msg}(u).\end{align}
The graph representation update at layer \(l\) is
\begin{align} g^{l} = g^{l-1} + \phi_{glb}\left( (g^{l-1},
  f(\vertex)) \right)\end{align} where \begin{align}
  f(\vertex) = \sum_{v\in\vertex} \text{softmax}(\phi_{att}(v))
  \cdot \phi_{feat}(v),\end{align}
an attention-based aggregation scheme proposed by~\cite{li2019graph}. Each function \(\phi\) is implemented as a single-layer neural network,
\(\phi(x; \mathbf{W}, b) = \rho(x\mathbf{W}^T+b)\), where \(\rho\) is a non-linear
activation function.

\subsection{Policy Decomposition}






The local approach decomposes the policy of the agent into two actions, where the agent first selects a node in the graph and then what action should be taken on it. In our notation,
\(E\) denotes the size of the representation vectors and \(N\) the amount of nodes, \(|\vertex|\).
\(\textbf{H}^l_{N\times E} = \{h_1^l, \ldots, h_{N}^l\}\) is a multiset of node
representations after \(l\) message passes. \(\mathcal{C}\) the set of commands from CAGE 2. We formulate the factorized policy function as
\begin{align}
  \pi(v_i, c_i | \nodeembed{}) = \pi_2(c_j|h_i^l, v_i)\pi_1(v_i|\nodeembed{})
\end{align}
where
\begin{align}
   & \nodepolicy{} = \sigma_i(\{\nn_1(h_1^l), \ldots, \nn_1(h_N^l)\}) \\
   & \condpi = \sigma_j(\nn_2(h_i^l))\end{align}
and
\begin{align} \sigma_i(x) =
  \frac{e^{x_i}}{\sum^{K}_{j=1}e^{x_j}},\end{align}
\(\nn_1 : \mathbb{R}^E \rightarrow \mathbb{R}\) and \(\nn_2 : \mathbb{R}^E
\rightarrow \mathbb{R}^{|\command|}  \).







\begin{figure}
    \centering
    \includegraphics[width=0.4\textwidth]{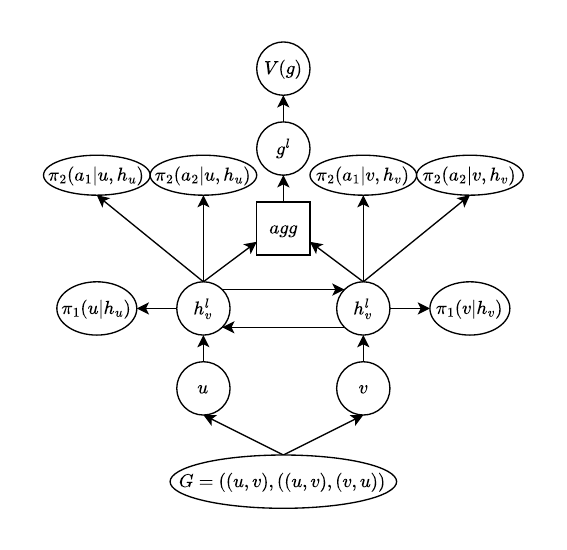}
    \caption{Schematic showing the computation paths for action probabilities and state value on a graph with two node and two edges.}
    \label{fig:graph}
\end{figure}

The value function estimation is defined as a function of the graph encoding
\(g^l_{1\times E}\),
\begin{align} V^{\pi}(\graphembed) = \nn_v{}(\graphembed) \end{align}
where \(\nn_v : \mathbb{R}^E \rightarrow \mathbb{R} \).
The critic is only used during training, so the graph representation is not required during inference. Each function \(\nn\) was implemented as a linear
transformation, where \(\nn(x; \param, b) = x\param^T +
b\). A schematic showing the computation paths of the various values is shown in \autoref{fig:graph}.

\subsection{Changes to CAGE 2}

We introduced two changes to the CAGE 2 implementation, which we based on commit
b5a71c4\footnote{\href{http://github.com/cage-challenge/CybORG/tree/b5a71c4cf156bf450a0bb3ffdb63fc0340a10bd4}{www.github.com/cage-challenge/CybORG}} of \ac{cyborg}. Our options were to either work from the version of \ac{cage} 2 in the most recent commit of \ac{cyborg}, or the older version used for the challenge. We opted for the more recent version, as we found it easier to work with.

\begin{figure}
  \centering
  \includegraphics[width=0.40\textwidth]{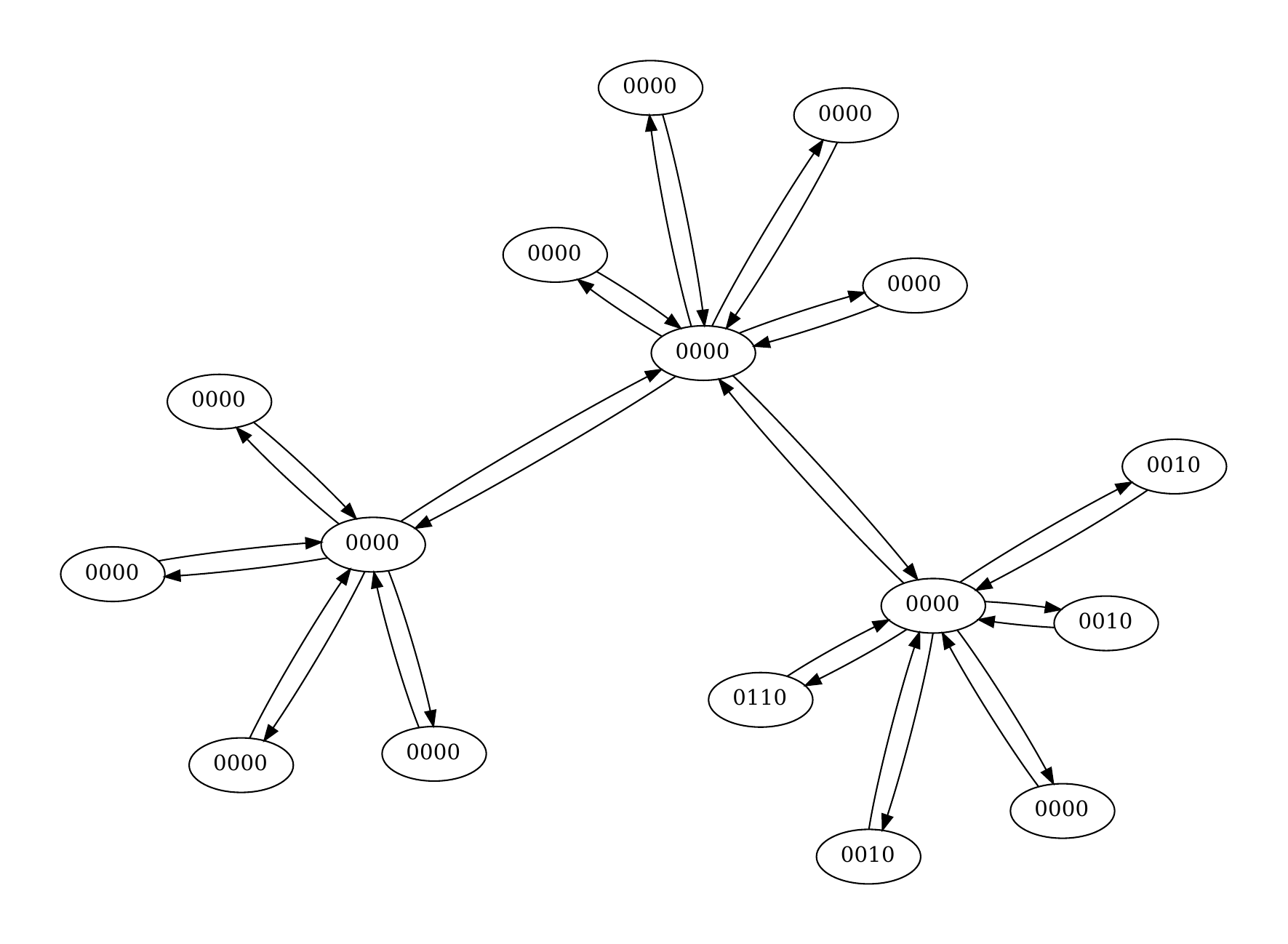}
  \caption{A state graph at a given point of the \ac{cage} 2 simulation. Each node has two
    categorical attributes encoded to binary using two bits each. The first
    attribute indicates the observed activity, and the second the access
    privilege of the red team agent on that host.}\label{fig:state-graph}

\end{figure}

The first change was reshaping the observation space to a graph, in order to use \iac{mpnn} agent.
We converted the existing \ac{cage} 2 vector state representation into a
homogeneous undirected graph. In the graph, each host is represented as a node 
with a single edge type relating them to other hosts. 
Each host has two attributes, same as in the original vector, as described in Section~\ref{sec:cyborg}.
Nodes were connected such that the graph matched the network layout described in the challenge description.
Routers were included as nodes, though no observations are associated with them in \ac{cage} 2. They were therefore set to always have no activity, never be compromised and not included in the action space of the blue team. An example of a state graph is shown in \autoref{fig:state-graph}.

The second change was to the set of allowed actions for the blue team.
We removed the host that the red team starts at from the set of hosts that the agent can interact with,
and decoy deployment actions were not used. Motivations for these changes are provided in Section~\ref{sec:changes}.

\section{Evaluation}
\newcommand{\experimentshuffle}{Host Reordering}
\newcommand{\experimentoriginal}{Full Network}
\newcommand{\experimentvariants}{Network Topology Changes}
\newcommand{\experimentlayers}{Message-Passing Steps}

Our evaluation of the \ac{mpnn} agent consisted of training agents in an environment configuration, 
and then testing the agent on a set of network variants. 
We only trained against the Meander agent, as it provided more varied
observations than the B-Line agent.
Previous work on \ac{cage} 2 trained against a combination of attackers, switching
between them at random between episodes~\cite{wolk2022cage}, or switched between models depending on the attacker~\cite{cyborgrewardshaping}. In our opinion, access to all forms of
red team behavior during training should not be assumed. We thus withheld the
B-Line policy during training and used it for evaluation only. 
The relational learning scheme we employ is not intended to generalize across opponent policies, 
and the ability to do so is only a fortunate side effect. 
Episodes were truncated after 50 steps during training. 
In the absence of blue team interference, this is enough for the
Meander agent to be able to reach the operational server.





To evaluate the ability of agents to generalize across structural changes,
we created variants of the original network by removing different hosts. 
Host were selected on the condition that red team agents using both the B-Line and Meander 
policies could reach the operational host without them. 
Given this condition, we generated six network variants, each one with an additional randomly selected eligible host removed. 
This resulted in seven network variants, which includes the original network of 16 hosts. 
Changing the number of hosts in the network not only changes the observation space of the problem, 
but also the action space, as the number of machines under the control of the blue team changes.
We trained \ac{mpnn} agents on the network variant with 13 hosts, and evaluate on all variants. 
This setup gave three larger networks, and three smaller, to evaluate agents on.
A specialized \ac{mlp} agent was trained for each network variant for comparison with the general \ac{mpnn} agent. We chose 
\iac{mlp} agent for comparison as they have been used in previous work on \ac{cage} 2~\cite{wolk2022cage}. 

Agents were rated by the scoring scheme described in Section~\ref{sec:cyborg}, that test agents with different episode lengths and red teams, as in previous work~\cite{wolk2022cage, cyborgrewardshaping}.

The scale of the rewards change depending on the network size and episode length, making comparisons of agents for multiple problem instances difficult.
As an alternative performance metric, we calculated the percentage of episodes where the agent did not receive \emph{any} negative reward, which we call a \enquote{perfect round}. 
A perfect round thus means that the blue team managed to stop the red team without receiving a penalty. 

Agent training and evaluation was done on a machine equipped with 
32 GB of RAM, an Intel Xeon Silver CPU with 24 1GHz cores and an NVIDIA Quadro RTX 4000 GPU. Training is also possible on more limited hardware, such as a
laptop equipped with 16 GB of RAM and an 11th Gen. Intel i7 CPU. Our code implementation relies on PyTorch Geometric for the message-passing
algorithm. We used \acl{ppo} to train agents, as was also done by~\cite{wolk2022cage}. 
The \ac{ppo} and \ac{mlp} implementations were sourced from Stable Baselines 3.
The code for the \ac{mpnn} policy is based on the implementation
by~\cite{oracle-sage}, who in turn ported the code from
the original authors~\cite{janisch2023symbolic}
to work with Stable Baselines 3. 
Different selections of hyperparameters were not extensively evaluated, 
but we present results for \ac{mpnn} agents with different numbers of layers. 
We do not include \ac{mpnn} agents with a layer depth of one, as these were significantly worse than the layer variants. 
All agents were trained using 500\,000 steps of each environment in total.

\section{Evaluation Results}
\begin{figure*}
    \centering
    \includegraphics[width=\textwidth]{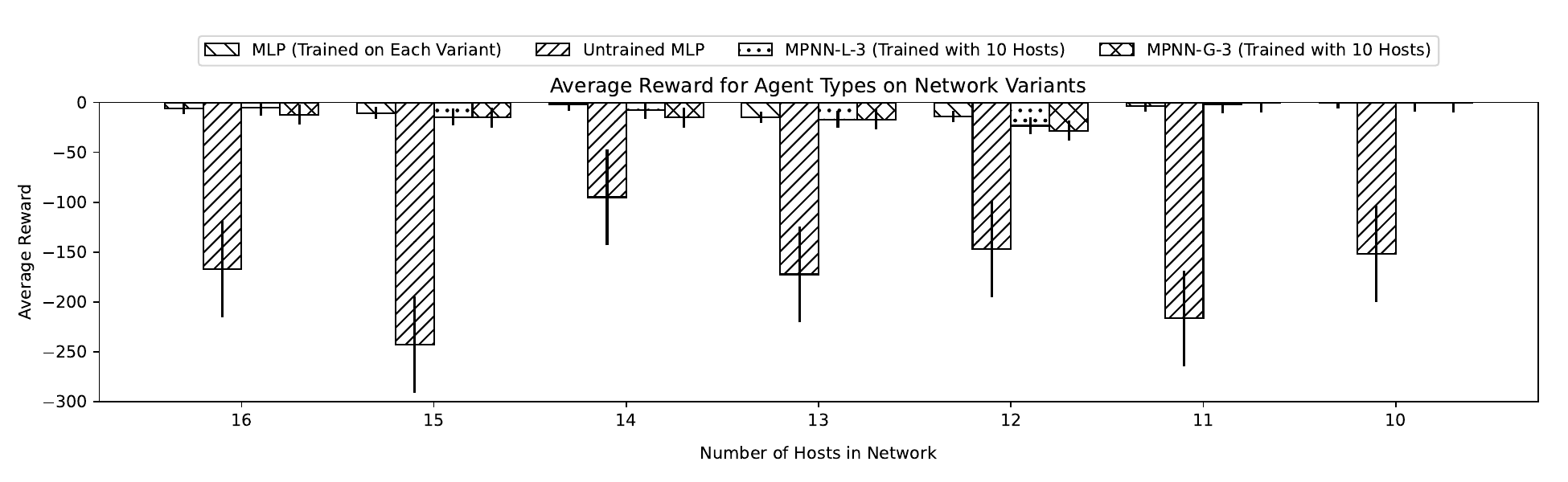}
    \caption{Bar chart showing rewards for trained and untrained agents averaged over 1000 episodes. MLP agents were specially trained on each network variant. MPNN agents were only trained on the network variant with 13 hosts, and evaluated on others. Agents were trained and evaluated against the Meander red team policy, with episode lengths of 50 steps.}\label{fig:bar-chart}
\end{figure*}\label{sec:results}

An immediate difference between agents using \ac{mlp} and \ac{mpnn} models was the
training time. With our implementation, \ac{mpnn} training was roughly 10
times slower than \ac{mlp} training. In concrete numbers, training each \ac{mlp} model took roughly 4 minutes, and each \ac{mpnn} model 40 minutes. The difference was alleviated, however, by the fact that we trained less \ac{mpnn} agents overall, as we did not have to train a separate agent for every scenario variant.


We calculated average rewards and the total score following the challenge rules,
meaning that each combination of attacker, episode length and agent was
evaluated for 1000 episodes. The total score presented in
\autoref{tab:variant_perfects_averages} is the sum of average rewards for each
configuration. Average rewards for each network variant is shown in \autoref{fig:bar-chart}. We denote the \ac{mpnn} variants with the shorthand
\enquote{[GL]-\(x\)}, 
where \enquote{G} and \enquote{L} denote if the model uses global or local representations described in Section~\ref{sec:message-passing},  and \(x\) denotes the number of layers the model uses. 
We emphasize that scores are not directly comparable to those
of previous work, such as~\cite{cyborgrewardshaping} or~\cite{wolk2022cage}. 
The version of CybORG is different, and we omitted actions from the action set that is to the benefit the blue team. 
We also note that our comparisons are primarily qualitative. 
The difference between the \ac{mpnn} models and the \ac{mlp} is not that the \acp{mpnn} performs better or worse at the different scenarios, 
it is that a trained \ac{mlp} agent simply \emph{can not function at all} without modification when the observation space changes in size.
In general, quantitative differences in rewards between agent variants are not 
easy to discern with \ac{cage} 2, as has been noted previously~\cite{wolk2022cage}. 
Deviations in scores is high for all agents, relative to the average. 
We include the scores of an untrained \ac{mlp} agent, which serve as a baseline zero-shot solution to the problem.

Our primary result is that \ac{mpnn} agents received scores and percentages of perfect rounds
significantly higher than the baseline on network variants not trained on.
Scores and the percentages of perfect rounds of \ac{mpnn} models were lower than those of the trained \ac{mlp} ensemble, but we note that this is comparing a single agent using one \ac{mpnn} to an agent using a collection of seven \ac{mlp} models 
specially trained on each network variant. 
L-3 received the highest total score out of the \ac{mpnn} variants, and L-4 the highest percentage of perfect rounds.
The local versions of the \ac{mpnn} have higher total scores than their global
equivalents in aggregate. This lends some credit, albeit not much given the high
variance, to our hypothesis that local information is sufficient to find a solution to this problem. 
The \ac{mpnn} agents obtain perfect scores on two network variants that the trained \ac{mlp} agent does not. 
The untrained \ac{mlp} agent ensemble never achieves a perfect round, indicating that this can not be done through random action. 
Table~\ref{tab:variant_perfects_averages} shows the total scores for all \ac{mpnn} variants and the \ac{mlp} ensemble agents, 
averaged over all seven network variants. \autoref{tab:perfectsfull} shows the percentages of perfects rounds with each network variant.

\begin{table}[!htpb]
\centering
\caption{Percentages of episodes with no penalties averaged over all red team and episode length configurations. \ac{mpnn} agents were only trained with 13 hosts. \ac{mlp} agents were trained on each network variant.}
\begin{tabular}{lrrrrrrr}
\toprule
Hosts & 16 & 15 & 14 & \textbf{13} & 12 & 11 & 10 \\
\midrule
MPNN & \multicolumn{7}{c}{Episodes With No Penalty (\%)}  \\
\midrule
G-2 & 5 & 0 & 17 & 0 & 0 & 32 & 0 \\
G-3 & 1 & 2 & 28 & 23 & 0 & 50 & 74 \\
G-4 & 0 & 0 & 1 & 0 & 0 & 7 & 1 \\
L-2 & 12 & 0 & 0 & 0 & 0 & 0 & 0 \\
L-3 & 17 & 0 & 32 & 18 & 0 & 8 & 49 \\
L-4 & 10 & 4 & 44 & 34 & 0 & 34 & 89 \\
\midrule
MLP & \multicolumn{7}{c}{Episodes With No Penalty (\%)}  \\
\midrule
Trained & 49 & 0 & 77 & 50 & 0 & 0 & 100 \\
Untrained & 0 & 0 & 0 & 0 & 0 & 0 & 0 \\
\bottomrule
\end{tabular}\label{tab:perfectsfull}
\end{table}
\begin{table}[!htpb]
\centering
\caption{Sum of average rewards and percentages of episodes with no penalties, aggregated over all combinations of network sizes, episode lengths and red team policies. \ac{mpnn} models
were only trained with 13 hosts.
\Ac{mlp} results are averages for seven agents trained on each network.}
\begin{tabular}{lrr}
\toprule
Agent & Episodes With No Penalty & Sum Average Reward \\
\midrule
MPNN & & \\
\midrule
G-2 & 8\% & -154 \(\pm\) 42 \\
G-3 & 25\% & -270 \(\pm\) 252\\
G-4 & 1\% & -172 \(\pm\) 63 \\
L-2 & 2\% & -136 \(\pm\) 22\\
L-3 & 18\% & -113 \(\pm\) 63 \\
L-4 & 31\% & -150 \(\pm\) 99\\
\midrule
\multicolumn{3}{l}{MLP (Ensemble)} \\
\midrule
Trained &  39\% & -78 \(\pm\) 42 \\
Untrained & 0\%  & -2104 \(\pm\) 605\\
\bottomrule
\end{tabular}\label{tab:variant_perfects_averages}
\end{table}

\section{Related Work}
There exists several cyber incident simulators with different
assumptions\footnote{\href{https://github.com/Limmen/awesome-rl-for-cybersecurity}{github.com/Limmen/awesome-rl-for-cybersecurity}}
and associated solution methods~\cite{ADAWADKAR2022105116}. In the interest of
space, we limit works in described in this section to those that use \ac{cage} 2
in some form. 
We make one exception to this limitation and highlight the work of
\cite{collyer2022acd}, that combines reinforcement learning, relational
graph learning and cyber security similarly to us. 
They evaluate their approach on another cyber incident simulator, Yawning Titan, which is more abstract than \ac{cyborg}.
Our implementation differs from theirs in that they use a graph embedding algorithm named Feather-G, a method for graph representation based on characteristic functions~\cite{feather}. 
It is not explicitly stated in the text that the action space is generalized to different number of hosts. 
In principle, we could use the node-level version of Feather, Feather-N, to represent nodes as we do with the \ac{mpnn}.

Our starting point for studying previous work on \ac{cage} 2 are agents that obtained high scores on the challenge.
The winners of the challenge, a team from Cardiff University, used
\iac{mlp} agent trained using \ac{ppo} in combination with handwritten heuristics. 
The agent uses a static policy for deploying decoys, derived from a greedy search over the space of possible decoys that can be deployed.
The agent also uses hard-coded heuristics to determine which one out of two
policies to use in a given episode, one for each red-team policy, and limit the blue team's actions to those that are strictly useful. 
Although effective in the challenge, this agent formulation can not generalize to unseen configurations, as its scenario-specific heuristics would need to be manually updated. The runner-up on the leaderboard is another \ac{mlp} agent by a team from the Alan Turing Institute, also trained using \ac{ppo}. 
It uses two policy functions with different observation spaces that are switched between by a higher-level policy.  
The policy used against Meander uses the unmodified observation space,
and the one used against B-Line adds information about the number of ports to each host.
In terms of the ability to generalize, this is better than the Cardiff approach as it does not hard-code parts of the policy. 
However, it is still bound to a particular size of the network.

There are works outside the challenge that have used \ac{cage} 2 for development and evaluation.
\cite{cyborgrewardshaping} trained agents with different forms of reward shaping on the
environment. They base their work on the model by Cardiff University, which in our opinion makes 
conclusions difficult due to the issues of hard-coding mentioned previously.
\cite{ZHU2024103578} trains agents on \ac{cage} 2 using a version of deep Q-learning. Similar to us, they create three scenario variants to test their agent, but have to train a separate agent on each version due to their approach being locked to a specific network size.
They also note that their approach has issues when introducing additional hosts to the network, but the causes of these issues are not made clear.
\cite{wolk2022cage} evaluates a set of neural network models on \ac{cage} 2
on different configurations. 
The experiments are intended to evaluate attributes that they consider important when
agents are applied in real systems, such as inference time and generalization. 
They note that their agents trained with \ac{rl} on \ac{cage} 2 are challenged by
changes to the network configuration, such as the topology, and differences in the attacker policy.








\section{Discussion}


From the results of our experiments, 
we see that the \ac{mpnn} agents perform better on the unseen network variants than the untrained \ac{mlp} baseline. 
We also see that they can obtain a nonzero percentage of rounds where the red team does not capture any machines, and the blue team does not receive any penalty.
We take this as evidence that the \ac{mpnn} agents are capable of zero-shot generalization in this problem domain. 
The zero-shot performance is worse than that of \ac{mlp} policies specifically trained on the network variant. This is in line
with results obtained by~\cite{janisch2023symbolic}, 
where the \ac{mpnn} agents tend to be outperformed by one-shot planners on specific problem instances. The difference in results indicates a trade-off between generalization and specialization. 
We do not know if a fully generalized policy is possible to find for this problem, but work has been done to analyze this question in other domains~\cite{DBLP:conf/nips/MaoLTK23}.
We chose \ac{mlp} agents as an upper bound of performance, as they are often used in previous work on the \ac{cage} 2 environment. 
However, these are likely worse than a theoretical optimal policy, meaning that we do not know how close the zero-shot policy is to optimal.
We can see from~\autoref{tab:perfectsfull} that for some network
variants, both the \ac{mlp} and \ac{mpnn} agents have few to no perfect rounds. 




\subsection{Limits of Relational Learning}

An inductive bias of relational reinforcement learning
is that the properties and relationships between objects remain constant.
This may not be true, depending on the problem domain.
In our problem, changing the red team policy will change the underlying dynamics 
of the problem, and the change in representation that the \ac{mpnn} 
introduces does not directly address this.
It may indirectly do so if the relational rules learned by the
agent is general enough to counter any red team policy, 
but this is also true for a policy that can be learned by \iac{mlp} agent.
Other methods of zero-shot learning would need to be applied to address this issue~\cite{DBLP:journals/jair/KirkZGR23}.

\subsection{Changes made to CAGE 2}\label{sec:changes}

We made a set of changes to the \ac{cage} 2 environment.
We removed the host that the red team starts at from the blue team action space. The starting host is different from other hosts in that blue team commands fail when used on it. This is to ensure the red team always has access to the network. 
If vector observations are used, there is no way to identify the starting host other than by its position in the vector.
We chose to not include decoy actions for similar reasons. The decoy actions have specific sets of requirements that need to be fulfilled in order to execute them. This includes ports not being in use,
or the host running a particular operating system. This information is not present in the wrapped 
observation space, and only the position of the host in the vector encodes this information.
In both these cases the position of the host in the state vector acts as an implicit 
host identifier. This information is lost when the vector is reshaped to a graph, but makes 
the agent using the graph more general. We can assign a unique label to each node in the graph, 
which would allow an agent to fully differentiate between objects. 
However, it also means that the rules learned will be specific to that set of objects and less generally applicable. We could also add the object class to nodes, but then we have to design a fitting ontology.
Our opinion is that \ac{cage} 2 is an interesting environment, albeit one that is slightly oversimplified.
We believe future work should focus on utilizing the unwrapped observation space, which contains information more in line with what an actual network intrusion detection system would produce.

\section{Conclusion}

We implemented agents for automated network intrusion response that use \aclp{mpnn} to encode facts about the network.
The agents were evaluated using a simulated network environment, the \acl{cage}. 
Our results show that our agents can generalize across network variants without additional training but are outperformed by specially trained policies, indicating a trade-off between generalization and specialization.
Our work addresses an issue present in previous work on automated incident response~\cite{wolk2022cage, cyborgrewardshaping}:
That agents are bound to a specific size, structure and ordering of the network topology. 
This is in spite of computer networks being highly variable in structure, and 
a network operator needs to handle such changes while enacting security policies.
We believe that by exploiting relational structure in the problem, 
agents for cyber incident response can be made more general and reusable.
Reusing agents in problems with a different structure, 
but similar dynamics saves both time and energy, which 
we consider important for practical use.





\bibliographystyle{IEEEtran}
\bibliography{refs}

\end{document}